# Chemical transformer compression for accelerating both training and inference of molecular modeling

Yi Yu and Karl Börjesson[*]

Department of Chemistry and Molecular Biology, University of Gothenburg, Kemivägen 10, 412 96 Gothenburg, Sweden.

[*] Author to whom any correspondence should be addressed.

E-mail: karl.borjesson@gu.se

**Keywords:** model compression, transformer, molecular modeling

**Abstract:** Transformer models have been developed in molecular science with excellent performance in applications including quantitative structure-activity relationship (QSAR) and virtual screening (VS). Compared with other types of models, however, they are large, which results in a high hardware requirement to abridge time for both training and inference processes. In this work, cross-layer parameter sharing (CLPS), and knowledge distillation (KD) are used to reduce the sizes of transformers in molecular science. Both methods not only have competitive QSAR predictive performance as compared to the original BERT model, but also are more parameter efficient. Furthermore, by integrating CLPS and KD into a two-state chemical network, we introduce a new **de**ep **li**te **c**hemical **t**ransform**e**r model, DeLiCaTe. DeLiCaTe captures general-domains as well as task-specific knowledge, which lead to a **4x** faster rate of both training and inference due to a **10**- and 3-times reduction of the number of parameters and layers, respectively. Meanwhile, it achieves comparable performance in QSAR and VS modeling. Moreover, we anticipate that the model compression strategy provides a pathway to the creation of effective generative transformer models for organic drug and material design.

## 1. Introduction

By silico modeling and analysis of chemical structures, molecular computational approaches have facilitated the development in various fields, such as drug discovery and material design [1-3]. Nowadays, deep learning methods have made significant breakthroughs in these fields [4-6]. To go from a chemical structure to a computational descriptor, molecules are encoded by different representations (figure 1(A)) [7], such as the Simplified Molecular Input Line Entry Specification (SMILES) [8] or 2D undirected cyclic graphs [9]. Then, a suitable neural network is designed and trained to connect the molecular representation to chemical structure for discriminative or generative tasks. It is essential to introduce a suitable model that are optimize for one's purposes. On the one hand, discriminative models could be utilized for predicting physicochemical properties, reaction performance and bioactivity [10]. One the other hand, generative models could design novel molecules efficiently and automatically with respect to specific objectives [11-13].

In the last few years, transformer models have shown to be an efficient deep learning method within chemical science [14]. They have formed a new paradigm and works by self-supervised learning as the pre-training step. This self-supervised learning is based on large unlabeled chemical sequence or graph datasets. Then the models are fine-tuned in downstream tasks [15]. More concretely, models could learn general-domain chemical knowledge (e.g. the understanding of the correct valency of chemical bonds etc.) in the pre-training process. Then this knowledge is transferred in the fine-tuning to overcome data scarcity for specific tasks, such as quantitative structure–activity relationship (QSAR) [16-20], virtual screening (VS) [21], reaction prediction [13], de novo design [15], molecular optimization [22] and drug-drug interaction predictions [23].

Despite being high performance, given the size of at least several ten million parameters, transformer models are computationally expensive in both training and inference processes [24]. For instance, recently proposed models require several days of pre-training with at least 4 high performance GPUs [15, 23]. Furthermore, the inference time in the fine-tuning process is much longer than in other models [25]. Besides being time consuming, the environmental cost of the hardware running should be considered as another disadvantage. Hence, the growing computational and hardware requirements is likely to hinder the wide practical use of transformer models [26].

In order to improve the parameter-efficiency, model compression has recently been an active research area in the fields of computer vision (CV) and natural language processing (NLP). By compression, large models, like transformers, could be scaled down and the training or inference processes are then accelerated. Cross-layer parameter sharing (CLPS) is one strategy to decrease the computational cost of the training process, while retaining the high performance of downstream tasks [26, 27]. In CLPS, the default decision is to share all parameters across layers. In other words, parameters are tied across positions and time steps. Additionally, knowledge distillation (KD) is reported to result in lighter models with faster inference [25, 28]. In KD, a student model with small size is trained to reproduce the behavior of the teacher model. Nonetheless, neither of them has been utilized in transformer models within chemical science.

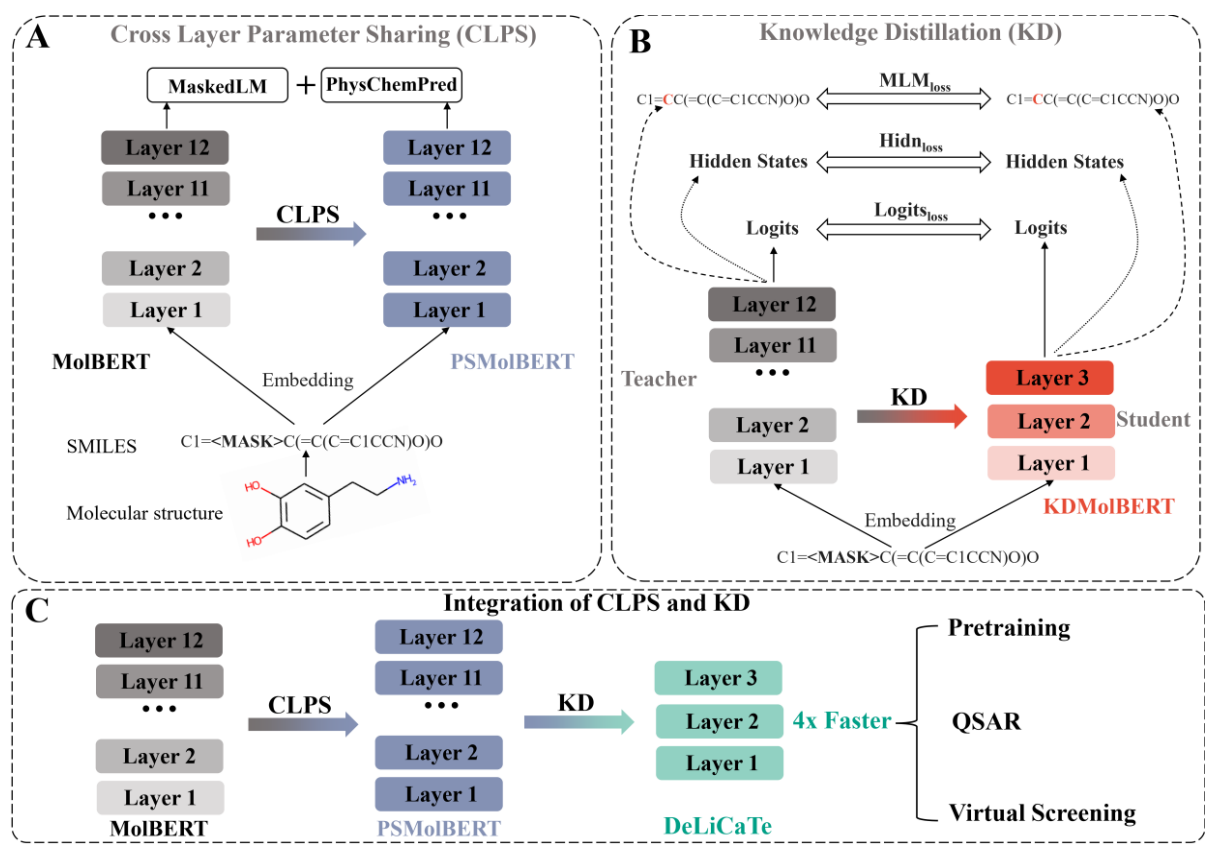

**Figure 1.** An illustration of the processes of the CLPS, KD methods, as well as their implementation with MolBERT forming DeLiCaTe. **(A)** Molecule are encoded by SMILES representation. Next, the PSMolBERT is pretrained based on SMILES which passes into embedding and 12 parameter-sharing attention layers in turn. Then, the loss is calculated by MaskedLM and PhysChemPred as two self-supervised tasks [21]. The same color depth of layers in PSMolBERT represents the parameter sharing feature in-between layers of CLPS, while the different color depth symbolizes that the parameters are different among layers in MolBERT. **(B)** The KDMolBERT is obtained by general distillation in which the pretrained MolBERT is acted as a teacher model. The student model, KDMolBERT, is distilled to reproduce the behavior of the teacher model, the pretrained MolBERT. The behavior is evaluated by triple losses, including logits, hidden states and MaskedLM. **(C)** The DeLiCaTe model integrates the advantages of the two mentioned method. It is distilled from the pretrained PSMolBERT, and a 4x speedup is realized in both training and inference compared with MolBERT.

Inspired by model compression within CV and NLP applications [29-34], we aim to explore more parameter-efficient transformer models for molecular modeling based on SMILES-based molecular representation. The concept is illustrated in figure 1. We begin by using CLPS or KD strategies independently to compress a high-performance chemical transformer, MolBERT. Then the training and inference time is compared among the models, as well as the model sizes. Next, we show that QSAR performances of the two compressed models are competitive on various tasks, compared with MolBERT. Finally, given the positive effects above, the CLPS and KD are integrated with each other to establish a new **de**ep **li**te **c**hemic**a**l **t**ransform**e**r model, DeLiCaTe (shown in figure 1(C)). DeLiCaTe achieves more than 96% of the performance of MolBERT in QSAR as well as VS applications, while being **4x** faster in the rates of training and inference due to a ~**10-** and **3**-times reduction of the number of parameters and layers, respectively.

## 2. Methods

### 2.1 Dataset for Model Compression

We used the GuacaMol benchmark dataset [11] from ChEMBL [35] containing ~1.6 M compounds for pre-training or distillation. The setting was used for the pre-training dataset in MolBERT as well [21]. For the dataset, the ratio of training to validation was 16:1.

### 2.2 Cross-Layer Parameter Sharing

MolBERT is a bidirectional chemical model derived from the BERT architecture [21]. It is a well-recognized molecular transformer with state-of-the-art QSAR and VS performance. The backbone of the CLPS model (PSMolBERT) uses a transformer encoder with gelu activation function, similar to MolBERT. The default decision is to share all the parameters across the layers. In other words, the weights among different layers are same. The method for pre-training PSMolBERT is shown in figure 1(A). The pre-training tasks include masked language modeling (MaskedLM) and calculated molecular descriptor prediction (PhysChemPred), which is consistent with the original MolBERT [21]. The first one is the similar self-supervised learning method to NLP. And the latter one is to predict the normalized set of chemical descriptors for each molecule. The general-domain chemical knowledge is well-learned by these two tasks. Following this method, the parameter-efficiency of CLPS is studied by directly comparing the modeling performance with MolBERT and other baseline methods.

### 2.3 Knowledge Distillation

Regarding the mechanism of KD, as shown in figure 1(B), a student model is trained to reproduce the behavior of the teacher model. The behavior is evaluated by the loss function which is a linear combination of 3 types of losses:

$$L = L_{mlm} + L_{hidn} + L_{logits} \qquad (1)$$

Where L is the final training loss, $L_{mlm}$ is the masked language modeling loss in the supervised learning process, and $L_{hidn}$ and $L_{logits}$ are the losses of hidden state and logits between the student and teacher model calculated by the mean square error. As for the loss of logits, following Hinton et al. [34], the softmax-temperature method was used where the temperature (T) controls the smoothness of the logits distribution and the same T was applied to the student and teacher in the training period. Here, the T is 8 for KD. Meanwhile, the ablation study for KD is conducted by removing any one of the triple losses.

The teacher model here was the pre-trained MolBERT with 12 transformer layers, while the student only had 3 layers of which the weights were randomly initialized. The other structures between these two models remained the same and are shown in Supplementary table S1.

### 2.4 the DeLiCaTe model

In order to obtain DeLiCaTe, MolBERT was first pre-trained to get the PSMolBERT by CLPS. Then, the PSMolBERT model was distilled to obtain DeLiCaTe (figure 1(C)). The detail parameter and architectures are shown in Section 3.1.

**2.5 Experiments, Baseline and Evaluation**

The performances of the aforementioned models were evaluated in QSAR and VS experiments. In addition, three baseline methods were used: i) RDKit descriptors [36], ii) Extended Connectivity Fingerprints with a diameter of 4 (ECFP4), and iii) CDDD [37], which is a recurrent neural network (RNN) model that today achieve state-of-art performance in QSAR and VS experiments.

Ten QSAR tasks were selected to compare the performance of the different models. The QSAR datasets were taken from MoleculeNet [38] and other sources in which one half of them are classification tasks and the other half are regression tasks. Table 1 summarizes the description of 10 datasets. The data splitting method followed the strategy provided from ChemBench in which the splitting ratio was 80% on training, 10% on validation, and 10% on testing. Then, a 3-fold cross validation was conducted. Finally, the area under the receiver operating characteristic (ROC-AUC) values and coefficients of determination ($r^2$) were used in the classification and regression tasks, respectively.

**Table 1.** Overview of the ten different QSAR datasets.

| Dataset | Acronym | Number of molecules | Reference | Type of tasks |
|---|---|---|---|---|
| Human β-secretase 1 inhibitors | BACE | 1483 | [38] | Classification |
| Blood–brain barrier penetration | BBBP | 1879 | [38] | Classification |
| Inhibition of HIV replication | HIV | 41101 | [38] | Classification |
| Mutagenicity | AMES | 6130 | [39] | Classification |
| Endocrine disruptors | Eds | 817 | [41] | Classification |
| Aqueous solubility | ESOL | 1128 | [38] | Regression |
| Free solvation energy | FreeSolv | 642 | [38] | Regression |
| Lipophilicity | Lipo | 4200 | [38] | Regression |
| Epidermal growth factor inhibition | EGFR | 4113 | [40] | Regression |
| Fibroblast growth factor receptor | FGFR1 | 4177 | [42] | Regression |

The VS was conducted on 69 datasets of which each one represents an individual protein target and contains a small number of active molecules amongst a much larger number of inactive ones. The benchmarking protocol by Riniker et al [43] was followed. Then, the ROC-AUC was used as the metrics to report the result.

**2.6 Implementation and Hardware**

The models were implemented by PyTorch [44] and Hugging Face Transformers [45], and TextBrewer [46] was adapted for chemical model distillation. Additionally, an Adam optimizer was used for both the pre-training and fine-tuning processes. The vocabulary size for all the chemical transformer models was 42. One NVIDIA RTX 3080 was used for all the training.

## 3. Results and Discussion

**3.1 Configurations and Speeds of Models**

The main differences among the chemical transformer models in this study are the number of parameters and attention layers. Table 1 outlines the used configurations and relative rates of models. As for the CLPS method, the number of parameters of the new model (PSMolBERT) was compressed into ~12% of the original one, which accelerates pre-training with 1.3x per epoch. Meanwhile, according to the validation loss curve (shown in Supplementary figure S1), the loss function of PSMolBERT converges much faster than MolBERT. This phenomenon has previously been observed within the NLP field, where parameter-sharing has an effect on stabilizing network parameters [27]. Hence, it is hypothesized that the pre-training could be significantly accelerated, not only in the rate per epoch but also by a fewer number of epochs for convergence. However, even though the number of parameters is tremendously decreased, the inference is not speeded up, which is consistent with previous work in NLP [26, 27]. As for the KD method, the 3-layer KDMolBERT was distilled from the original 12-layer MolBERT, while other settings were left unchanged. Given the significantly reduced transformer layers, a 3.8x speedup on inference was achieved. DeLiCaTe was distilled from PSMolBERT. It is ~4x faster on both training and inference compared with the original model. The reduced number of parameters obtained by CLPS and layers obtained by KD significantly speeds up the pre-training and inference processes, respectively. In other words, it combines the advantages of CLPS and KD. In the next sections we will discuss the performance of achieved models, in relation to QSAR and VS tasks.

**Table 2.** The configurations and training/inference time of the chemical transformer models.

| Model | Parameters (M) | Training/KD time[a] (h) | Time/epoch (h) | Layers | Inference speedup |
|---|---|---|---|---|---|
| MolBERT | 86 | 192 (100)[b] | 1.92 | 12 | 1.0 |
| PSMolBERT | 8 | 44 (30)[b] | 1.47 | 12 | 1.0x |
| KDMolBERT | 21 | 12.5 (13)[b] | 0.96 | 3 | 3.8x |
| DeLiCaTe | 8 | 4 (6)[b] | 0.67 | 3 | 3.8x |

Note: [a]The value given for MolBERT and PSMolBERT relates to the pre-training time, whereas the value given for KDMolBERT and DeLiCaTe represent the KD time. [b]The values in parentheses represent the number of epochs for pre-training or KD. The epoch training time was calculated as the ratio between the training time and number of epochs.

**3.2 Effect of CLPS on model performance**

3.2.1 Retained QSAR performance with CLPS

Having settled the model size and rate of training using the CLPS method, we now turn our attention to the performance. This was done by comparing the predictive capacity in QSAR of PSMolBERT with MolBERT as well as with other baseline methods, including RDKit descriptors, ECFP4 and CDDD [37]. Table 3 and 4 outline the results for classification and regression tasks, respectively. As for classification tasks, MolBERT pre-trained with 100-epoch performs the best among all models, which is accordant with the results from previous work [21]. Meanwhile, PSMolBERT pre-training with 30-epoch is not far behind MolBERT and outperforms traditional methods. Quantitatively, it retains 98.5% of the performance of MolBERT.

**Table 3.** The area under the receiver characteristic curve (ROC-AUC) for classification datasets.

| Method[a] | BACE | BBBP | HIV | AMES | EDC | Avg |
|---|---|---|---|---|---|---|
| RDKit | 0.844 | 0.757 | 0.776 | 0.801 | 0.853 | 0.807 |
| ECFP4 | 0.855 | 0.749 | 0.768 | 0.783 | 0.836 | 0.798 |
| CDDD | 0.832 | 0.823 | 0.771 | 0.807 | 0.872 | 0.821 |
| MolBERT | 0.907 | 0.910 | 0.830 | 0.894 | 0.941 | 0.896 |
| PSMolBERT | 0.891 | 0.904 | 0.823 | 0.879 | 0.917 | 0.883 |

Note: [a]The standard mean errors are shown in Supplementary table S2.

Regarding regression tasks, the trends are very similar to the ones for classification tasks. PSMolBERT is able to compete with MolBERT for regression modeling, achieving 99.2% of the performance of MolBERT on average.

In addition, the only difference compared to classification tasks is that the performance of CDDD now is comparable to two chemical transformer models. Among the individual tasks, CDDD performs the best in ESOL, FreeSov and Lipop, while two transformer models do better in EGFR and FGFR1. It indicates that CDDD could extract physicochemical knowledge from SMILES more efficiently, while transformer-based methods could perform better on bioactivity tasks, which is also found in previous literature [47]. However, averaging both classification and regression tasks, the two transformer models outperform all the traditional ones.

**Table 4.** Coefficient of determination ($r^2$) for regression datasets.

| Method[a] | ESOL | FreeSov | Lipop | EGFR | FGFR1 | Avg |
|---|---|---|---|---|---|---|
| RDKit | 0.870 | 0.795 | 0.729 | 0.659 | 0.666 | 0.744 |
| ECFP4 | 0.843 | 0.738 | 0.738 | 0.625 | 0.641 | 0.726 |
| CDDD | 0.920 | 0.834 | 0.797 | 0.671 | 0.715 | 0.787 |
| MolBERT | 0.905 | 0.816 | 0.780 | 0.720 | 0.730 | 0.790 |
| PSMolBERT | 0.892 | 0.832 | 0.758 | 0.710 | 0.721 | 0.784 |

Note: [a]The standard mean errors are shown in Supplementary table S3.

3.2.2 The effect of number of pre-training epochs on QSAR performance

According to the results of QSAR modeling, PSMolBERT achieves competitive performance and is 4.4x faster (in table 2, 192 / 44) on pre-training compared with MolBERT. Besides less training time per epoch, we assume that a quicker convergence contributes to the pre-training acceleration. We will therefore examine the QSAR performance as a function of pre-training epochs in more detail. PSMolBERT and MolBERT pre-trained with 10, 30, 60 and 100 epochs were compared with each other. Figure 2 displays the results of regression task performance. It demonstrates that the performance variation trends differ between PSMolBERT and MolBERT. For MolBERT, the performance improves as the pre-training epoch increases. However, in the case of PSMolBERT, no significant improvement is achieved after 30-epoch training. In other words, the best performance with respect to computational cost is achieved when the model is trained with 30 epochs. The result strengthens our hypothesis in Section 3.1 that PSMolBERT converges much faster than MolBERT, which lead to less pre-training time.

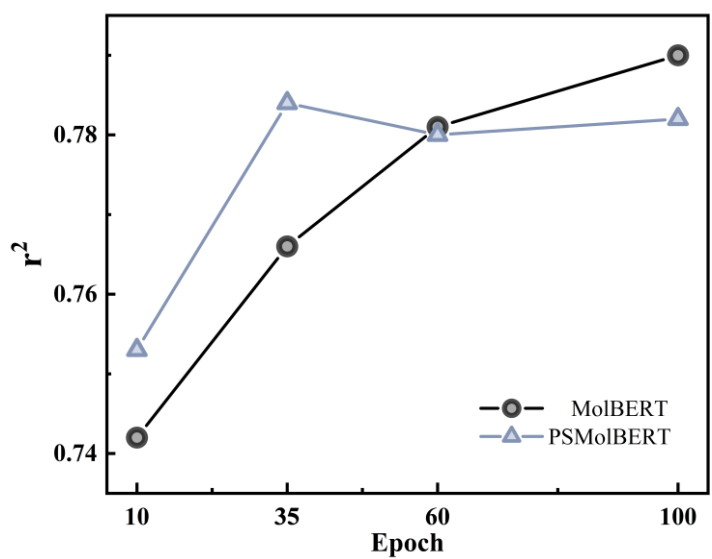

**Figure 2.** The QSAR performance of MolBERT and PSMolBERT as a function of pre-training epochs

Summing up, by CLPS, a 4.4x faster training speed is achieved while retaining about 99% of the performance of the original model. Therefore, CLPS can be concluded to be a parameter-efficient method for molecular modeling, reducing the high hardware requirements for pre-training transformer models. In the next section, we will discuss the effect of KD on inference speed and modeling performance.

### 3.3 Effect of KD on model performance

3.3.1 Retained QSAR performance by KD

We concluded in Section 3.1 that distillation resulted in a faster rate of inference due to less attention layers. We will now explore the effect of KD on the QSAR performance as well. Model distillation contains two lines of strategies, general- and task specific distillation [25]. General distillation is conducted by self-supervised learning to get general-domain knowledge, while task specific distillation compress models for specific tasks. In this work, only general distillation is studied on the chemical transformer models. By general distillation, the chemical knowledge is distilled from a teacher model to the student one using unlabeled data. The teacher model here is MolBERT, while the student model, KDMolBERT, is obtained after KD for 12.5 hours. Additionally, one random initialized 3-layer MolBERT model was pre-trained using the same time (12.5 h), named MolBERT-3. It was used as a baseline for comparison. Table 5 shows the results of the QSAR performance of classification and regression tasks on average. The results demonstrate that: 1) 99.3% and 96.7% of the performances of MolBERT were achieved by KDMolBERT in classification and regression tasks, respectively; 2) Except for CDDD on regression tasks, KDMolBERT outperforms other traditional methods in QSAR prediction (the performance by CDDD method is shown in table 4 and 5); 3) Even though the same time was used, the performance of KDMolBERT is much better than for MolBERT-3. These results suggest that the general-domain chemical knowledge can be effectively transferred from a teacher to student model in the KD process. Then, the compressed student model could be used for downstream tasks with 3.8x faster inference (shown in table 2).

**Table 5.** The QSAR performance of models by KD (KDMolBERT) and learning from scratch (MolBERT-3).

| Method | KDMolBERT | KD/MolBERT[a] (%) | MolBERT-3 | S/MolBERT[b] (%) |
|---|---|---|---|---|
| Classification | 0.890 | 99.3 | 0.821 | 91.6 |
| Regression | 0.764 | 96.7 | 0.714 | 90.4 |

Note: [a]The performance ratio between the student (KDMolBERT) and teacher model (MolBERT); [b]the performance ratio between 3-layer model learning from scratch and MolBERT.

3.3.2 Ablation study on distillation objectives

The loss function of the KD, also called distillation objective, includes the losses of masked language modeling, hidden states and logits (see Methods). The influence of each component in the triple loss was investigated by an ablation study. Table 6 presents the performance of removing each learning procedure. At first, the performances without logits significantly decrease by 3.6 and 4.1 percent units for classification and regression, respectively. The reason for the significant decrease lies in the mechanism of the QSAR modeling, in which pooled output from chemical transformers is used for classification or regression [21]. Both logits and pooled output is derived from the sequence output. When considering logits loss, the distribution of sequence output from the student model is matched with the one from teacher model. In order to receive high performance in fine-tuning processes, the logit loss need to be taken into account. Then, the impact of hidden states loss on performance was tested. The decreases of performance are 1.5 and 1.7 percent units for classification and regression, respectively. Therefore, the Effect were moderate, which is consistent with cases in NLP, and therefore not examined further. Lastly, the effect of

MaskedLM loss was examined, showing only minor changes in QSAR performance. Previous work on MolBERT indicated that additive gain from the MaskedLM is relatively minor, compared with the one in NLP. Our results indicate that MaskedLM has the least effect on chemical KD as well. Hence, the result above implies that the influence of each part in the triple loss is accordant with the one in NLP [24].

Table 6. Ablation study and variations to the model trained with triple loss.

| Without | Variation on QSAR | |
| --- | --- | --- |
| | Classification | Regression |
| $L_{logits}$ | -3.6 | -4.1 |
| $L_{hidn}$ | -1.5 | -1.7 |
| $L_{mlm}$ | -0.5 | -0.4 |

According to the abovementioned results, it can be concluded that KD is an empirically effective way to scale down the model size to facilitate inference and achieve competitive performance with its teacher chemical transformer. In the next section, the two compression methods, CLPS and KD, will be integrated with each other and the performance of the final model will be studied.

**3.4 Effect of Integration on QSAR performance**

The discussion so far has revolved around the effect of compression by CLPS and KD independently. These two compression methods work at separate stages in model constructions. CLPS reduces the time for training and KD for inference. They can therefore be applied within the same model. By applying CLPS on the original model MolBERT, we received PSMolBERT. Then by applying KD on PSMolBERT, we integrate CLPS and KD into one single model. We call this **de**ep **li**ght **c**hemical **t**ransformer DeLiCaTe, and will assess its performance in QSAR. The analysis of the QSAR performance of DeLiCaTe was done in an analogue's manner as in Section 3.3.1. As a baseline, a 3-layer PSMolBERT model (named PSMolBERT-3) was pre-trained from scratch to compare with DeLiCaTe. The pre-training time of the 3-layer PSMolBERT was the same as the distillation time of DeLiCaTe (6 h). Their QSAR performance is shown in table 7. DeLiCaTe achieves comparable performance to the teacher model, the 12-layer MolBERT. Specifically, it retains 97.2% and 94.7 of the performance of MolBERT for classification and regression tasks, respectively. It indicates that the combination of CLPS and KD not only compresses the model effectively, but also retains the chemical modeling ability. In comparison, the 3-layer PSMolBERT only achieve ~91% of the performance of MolBERT. It implies that a limited pre-training time and a small parameter scale (fewer number of layers) reduces the efficiency of molecular modeling by the learning from scratch method. Besides retaining performance, DeLiCaTe is ~4 faster on both training and inference compared to MolBERT (in table 2). Therefore, the integration of the two compression methods is an efficient strategy to counteract the time consuming and high hardware requirement of chemical transformers.

Table 7. QSAR performance of DeLiCaTe and 3-layer PSMolBERT in comparison with MolBERT.

| Method | DeLiCaTe | KD/MolBERT (%)[a] | PSMolBERT-3 | S/MolBERT (%)[b] |
| --- | --- | --- | --- | --- |
| Classification | 0.870 | 97.2 | 0.815 | 91.0 |
| Regression | 0.748 | 94.7 | 0.721 | 91.2 |

Note: [a]The performance ratio between DeLiCaTe (obtained by KD from PSMolBERT) and MolBERT; [b]the performance ratio between 3-layer PSMolBERT learning from scratch and MolBERT.

**3.5 VS performance of compressed models**

To further assess the efficiency of compression methods, the VS performance was studied with the aforementioned models. Given the compressed structure, the speed of VS was about 4x faster for DeLiCaTe than for MolBERT. Table 8 illustrated the average result of VS performance, and the performances of individual dataset are shown in Supplementary figure S2. The results demonstrate that all the compressed models achieve comparable performance. For instance, DeLiCaTe retain 96.8% of the performance of the original MolBERT. Considering the mechanism of VS, the results suggest that DeLiCaTe can discriminate the structural similarity among different molecules to facilitate the VS process. To strengthen this hypothesis, we further calculated the average pairwise cosine similarity of the molecules in the ChEMBL dataset using the DeLiCaTe, MolBERT and ECFP4 methods. As shown in Supplementary figure S3, the results of DeLiCaTe and MolBERT are very closed to each other and much lower than the one of ECFP4, which implies the excellent ability of molecular similarity discrimination by DeLiCaTe. Summing up, DeLiCaTe exhibits competitive performance with the original transformer and outperforms some of the baseline methods for VS tasks.

**Table 8.** AUC-ROC and standard deviation (SD) for Virtual Screening by methods mentioned above.

|         | RDKit | ECFP4 | CDDD  | MolBERT | PSMolBERT | KDMolBERT | DeLiCaTe |
|---------|-------|-------|-------|---------|-----------|-----------|----------|
| AUC-ROC | 0.633 | 0.603 | 0.725 | 0.743   | 0.737     | 0.730     | 0.719    |
| SD      | 0.027 | 0.056 | 0.057 | 0.062   | 0.059     | 0.066     | 0.064    |

**4. Conclusion**

In this work, we demonstrate the effect of implementing the CLPS and KD methods individually, as well as the integration of these two methods to compress chemical transformer. Both 4x speedup for training and inference were achieved by applying CLPS and KD, respectively. Furthermore, a deep light chemical transformer model, DeLiCaTe, was introduced to integrate the accelerating abilities of both compression strategies. According to the result of QSAR and VS performance, all the compressed transformers retain the molecular modeling capability of the original model and outperform or compete with state-of-the-art baseline methods. Consequently, time consuming and high hardware requirement are mitigated by the compressed parameter-efficient method. These results can facilitate the application of molecular discrimination based on chemical transformer encoders to a broader scientific community. Furthermore, due to the similarity between transformer encoders and decoders, this strategy is anticipated to promote the use of generative transformer models for organic drug and material design in the future.

**Data availability statement**

The data that support the findings of this study are available upon reasonable request from the authors.

**Acknowledgements**

We gratefully acknowledge financial support from the European Research council (ERC2017-StG-757733). Chao Fang is acknowledged for discussion on theory.

**Code availability**

All the code for model training and inference is available on https://github.com/YiYuDL/DeLiCaTe.

## Conflict of interest

The authors declare no competing interests.

## ORCID IDS

Yi Yu https://orcid.org/0000-0001-8360-005X
Karl Börjesson https://orcid.org/0000-0001-8533-201X